\documentclass{article}

\PassOptionsToPackage{numbers}{natbib}

\usepackage[final]{neurips_2019}

\usepackage{neurips_2019}
\usepackage{xcolor}

\usepackage[utf8]{inputenc} 
\usepackage[T1]{fontenc}    
\usepackage{hyperref}       
\usepackage{url}            
\usepackage{amsmath}
\usepackage{booktabs}       
\usepackage{amsfonts}       
\usepackage{nicefrac}       
\usepackage{microtype}      
\usepackage[pdftex]{graphicx}
\usepackage{caption}

\title{Classification as Decoder: Trading Flexibility for Control in Medical Dialogue}

\author{
    \textbf{Sam Shleifer}$^{1}$ \qquad
    \textbf{Manish Chablani}$^{2}$ \qquad \\
    \textbf{Anitha Kannan}$^{2}$ \qquad 
    \textbf{Namit Katariya}$^{2}$ \qquad
    \textbf{Xavier Amatriain}$^{2}$ \qquad \\
    $^{1}$ \text{Stanford, sshleifer@gmail.com}\qquad
    $^{2}$ \text{Curai, <firstname>@curai.com} \\
}

\begin{document}
\maketitle
\begin{abstract}
Generative seq2seq dialogue systems are trained to predict the next word in 
dialogues that have already occurred. They can learn from large unlabeled conversation datasets, build a deeper understanding of conversational context, and generate a wide variety of responses. This flexibility comes at the cost of control, a concerning tradeoff in doctor/patient interactions. Inaccuracies, typos, or undesirable content in the training data will be reproduced by the model at inference time.
We trade a small amount of labeling effort and some loss of response variety in exchange for quality control. More specifically, a pretrained language model encodes the conversational context, and we finetune a classification head to map an encoded conversational context to a response class, where each class is a noisily labeled group of interchangeable responses. 
Experts can update these exemplar responses over time as best practices change without retraining the classifier or invalidating old training data.
Expert evaluation of 775 unseen doctor/patient conversations shows that only 12\% of the discriminative model's responses are worse than the what the doctor ended up writing, compared to 18\% for the generative model.

\end{abstract}

\section{Introduction}

Modern chatbots are built around two paradigms: the more structured slot-filling paradigm, and the unstructured generative seq2seq paradigm. 

The first task-oriented group, exemplified by \cite{multiwoz}, tend to solve narrow tasks like restaurant and hotel reservations and require access to a large data structure.  
This setup is too cumbersome for primary care medical conversations because (a) building the external knowledge base would require the enumeration of the very large symptom, diagnosis and remedy spaces and (b) each module requires separate training data in large volumes.
The seq2seq group, which we call generative models (GM) require neither labeling nor structured representations of the dialogue state, but manage to learn strong representations of the conversational context with similar content to a knowledge base, according to \citep{fair}.
They have a key drawback, however: there are no mechanisms to ensure high quality responses. 
\cite{see} find that GM "often repeat or contradict previous statements" and produce generic, boring text, and GM can be attacked to "spew racist output" \citep{triggers}.
Even in a cooperative setting, typos, inaccuracies, and other frequent mistakes in the training data will be reproduced by the model at inference time. This drawback is even more important in medical settings, where giving patients bad advice is costly and potentially unsafe.

Our discriminative setup attempts to remedy this issue by restricting the chatbot to a manageable set of high quality “exemplar” responses. We ensure that exemplars are all factual, sensical and grammatical by allowing experts to edit them before or after training. For example, if we were to update a response recommending users sleep 6-8 hours per night to recommending 7-9 hours, we could simply update the message associated with the output class and the discriminative model would immediately generate the new advice in the same context it generated the old advice, without retraining. 

We address a key difficulty in this setup -- creating non-overlapping response groups that cover a wide range of situations -- with weak supervision. A pretrained similarity model merges nearly identical responses into clusters, and a human merges the most frequently occurring of these clusters into larger response classes. This results in a system that leverages novel pretraining techniques to generate useful responses in a wide variety of contexts, while still restricting generations to a fixed set of high quality responses.


\section{Related Work}
\label{sec:Related}
\par \noindent
\textbf{Healthcare dialog models:}
Most published dialog models in healthcare generate templated content supported by a knowledge graph ({\it c.f.} \cite{lit_review} for a comprehensive survey). \cite{fitz} proposes Woebot, a conversational agent designed to deliver cognitive behavioral therapy in the form of brief conversations with users. Underlying Woebot is a decision tree, where each node has a piece of content to send to the user, and (for some nodes) a proprietary NLP system, to determine which node to send the user to based on their most recent reply. \cite{minutolo2017conversational} prototypes a system for turning medical factoid questions into structured queries over a knowledge graph. The system covers a few example medical conditions, and asks the patient for more information until enough slots are filled to execute a valid query. Patient utterances must match a specific set of templates and synomyms to ensure correct queries.

Our work departs from this stream along multiple dimensions. First, we do not assume access to an external knowledge base. Second, we cover a wider range of medical conditions. Third, our model does not require perfect user input to generate good responses.
%


\textbf{Generative Dialog Models:} \cite{hface} won the 2019 PersonaChat competition with "TransferTransfo", a generative transformer approach.
The model starts training with pretrained weights from the GPT2 transformer, then finetunes with the PersonaChat data on a combination of two loss functions: next-utterance classification loss and language modeling (next word prediction) loss. 
Generation is performed in a typical generative manner: beam search with sampling and a blacklist to prevent copying from old utterances. 
We compare our architecture to this approach in Section \ref{sec: Experiments}.

\textbf{Discriminative Dialog Models:} The closest work to ours is \cite{airbnb}, AirBNB's customer service chatbot, which also uses a discriminative approach, but does not attempt to cover the whole response space and differs architecturally. Whereas our approach restricts the output space to 187 responses that attempt to cover the whole output space, the AirBNB system chooses from 71 one sentence investigative questions, each representing a cluster of questions, and leaves other response types, like statements and courtesy questions, to a separate model. 

\section{Approach}
\label{sec:Approach}

We aim to use the last $t$ turns of conversational context to suggest a response for a doctor to send to a patient. Our process involves two stages: (1) create groups of interchangeable doctor utterances, to use as labels for (2) train a classifier to predict a response class given the context that preceded it. 


\textbf{Weak Supervision Procedure for Generating Response Classes}
We aim to generate response classes, where each response class is a group of interchangeable responses observed in the conversation data, with the following characteristics: (1)   Low overlap between classes, (2) sufficient train examples in each class, (3) classes that cover a large number of unique responses. In the Figure below, we show our five stage procedure and briefly detail the steps in the figure's caption. A more detailed, mathematical explanation of the same procedure is presented in the Appendix.
\begin{figure}[!htbp]
\begin{center}
\includegraphics[scale=0.2]{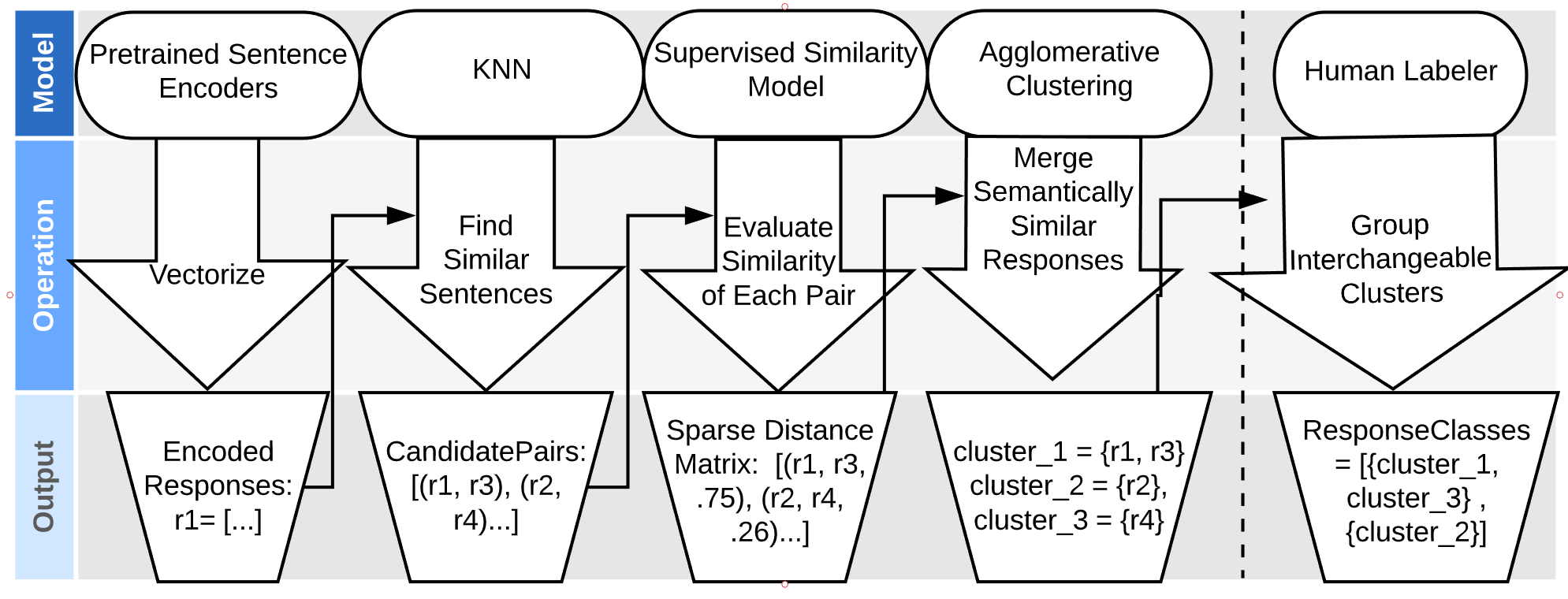}
\label{fig:response_pipe}
\end{center}
\caption{The leftmost steps (1) and (2) are designed to generate candidate pairs of responses in a semantic neighborhood, in order to avoid evaluating the similarity of O($\mathcal{N}^2)$ responses. We use BERT from  \cite{devlin2018bert} to evaluate similarity between  pairs of responses in step 3.  Step 4 runs Agglomerative Clustering the distance between two responses is their predicted probability of dissimilarity or 1 if they were not generated as a candidate pair.  Merging requires complete linkage, which means that two clusters are only merged if all the responses in both clusters are >= 75\% similar to all responses in the other cluster.
Step 5 is manual, and somewhat more involved. First create a dataset containing (centroid text, \# occurences of cluster constituents), sorted by \# occurrences, 
where the centroid is the most frequently occurring response in the cluster.
For each row in the dataset, the labeler decides whether the cluster centroid text belongs in any existing response class.
If centroid response could be used interchangeably with centroids in an existing group (most of the time): add the cluster to the existing response class. 
Otherwise, create a new group with a memorable name.
}
\label{fig:response_pipe}
\end{figure}


\textbf{Conversation Context $\rightarrow$ Response Class: Classification Training with ULMFit}
We train our discriminative response suggestion model to classify conversational context to one of the 187 response classes. (Context, ResponseClass) pairs are only included in the \textit{labeled} training data if the true response is a member of one of the response groups created in the step above.  We follow  \cite{ulmfit}'s ULMFit approach with a few modifications, most notably adding Label Smoothing to the loss function. The appendix details all modifications, and diagrams the training and  inference pipelines.
\section{Experiments}
\label{sec: Experiments}
\textbf{Data} For language model finetuning, we use 300,000 doctor/patient interactions containing 1.8 million rounds of Doctor/patient exchanges, collected through a web and mobile application for primary care consultations. We use the most recent 100,000 interactions, which contain 500,000 rounds as input to the Response Class Generation process, which yields 72,981 (context, response) pairs for classification training. The number of turns per conversation  (mean 10.8, std: 7.85) and length of each turn(mean: 20.4, std: 21.8) varies widely. 

\textbf{Clustering Statistics:}
Preprocessing and filtering yielded 60,000 responses. Step 2 yielded 1 million candidate pairs for evaluation. Step 4 yielded 40,000 response clusters with many overlapping groups; the largest cluster is only 10 distinct responses. In step 5, one labeler created 187 groups from the 3,000 most frequently occurring clusters in 3 hours. This leaves 90\% of all responses in the data unlabeled.
One advantageous property of our approach is that the human merging step need not be fully completed. In other approaches, like \cite{airbnb}'s, where there are fewer, larger automatically generated clusters and a human iterates through them and removes heterogeneneous constituents, every response must be considered. This would have taken us 40 hours, if we extrapolate linearly.

We hypothesize that fully automating the clustering process is difficult because the pretrained sentence encoders used in our candidate generation step are misaligned with our merge criteria, which is more permissive than pure semantic similarity. For example, none of our pretrained sentence encoders produce ("You're welcome. Hoping for the best.", "Take care, my pleasure.") as a candidate pair.

\subsection{Evaluation criteria}
\label{eval_crit}
\textbf{Expert evaluations for end-to-end comparisons.} To compare discriminative and generative approaches, we construct a test set constructed of (conversation, response) pairs that are held out from training and validation data. Roughly 91\% of test data responses are unlabeled. We call a response unlabeled if it is not an exact duplicate of responses in our 187 response class clusters.

Given the low correlation of automated metrics such as BLEU score to human judgment of response quality reported in \cite{metrics}, a group of medical doctors evaluated the quality of generated responses on the test data.
For a given conversational context, evaluators compared the doctor response observed in the data to a model's suggested response.  Evaluators reported whether a model's response is either (a) equivalent to the observed response, (b) different but higher quality, (c) different but equal quality, or (d) different but lower quality. For example, “Hoping for the best, take care.” and "Take care!!!!” would be marked equivalent. 

\textbf{Accuracy} on unseen \textit{labeled} data is used to compare different classifiers on the same dataset.

\subsection{Results}
We find that on 775 test set conversations, the discriminative model compares favorably to the generative model, generating responses evaluated as worse than those observed in the data only 12\% of the time, compared to 18\% for the generative model.
\begin{table}[!htbp]
\centering
\begin{tabular}{@{}lll@{}}
\toprule
                                & Generative & Discriminative \\ \midrule
a. Equivalent to Dr.                   & 56\%       & 71\%           \\
b. Different, higher quality & 1\%        & 6\%            \\
c. Different, equal quality  & 25\%        & 11\%           \\ 
d. Different, lower quality & 18\%        & \textbf{12\%}            \\ \bottomrule
\end{tabular}
\label{qresults}
\begin{tabular}{@{}lllll@{}}
\toprule
Architecture          & 4 Turn Accuracy & 8 Turn Accuracy & Encoder Finetune Time & Train Time \\
\midrule
ULMFit              & \textbf{56.70}\%         & \textbf{57.00}\%         & 12h                   & \textbf{40 mins}    \\
QRNN                & 49.30\%         & 49.20\%         & 0                     & 2h         \\
Hierarchical ULMFit & 53.80\%         & 54.90\%         & 12h                   & 18h        \\
Hierarchical QRNN $^\S$  & 47.80\%         & 49.40\%         & 0                     & 6h         \\
Transformer $^\dag$                & 56.64\%         & 56.82\%         & 12h                   & 6h         \\
\bottomrule
\end{tabular}

\caption{\textbf{Architecture comparison}: all experiments were completed on a single V100 GPU. $^\dag$ follows \cite{hface} and requires 10x slower inference than ULMFit. $^\S$ follows \cite{serban} and \cite{airbnb}. Details and discussion of the tradeoffs of different approaches can be found in the appendix. }
\label{archtab}
\end{table}

\begin{figure}[!htbp]
\begin{center}
\includegraphics[scale=0.2]{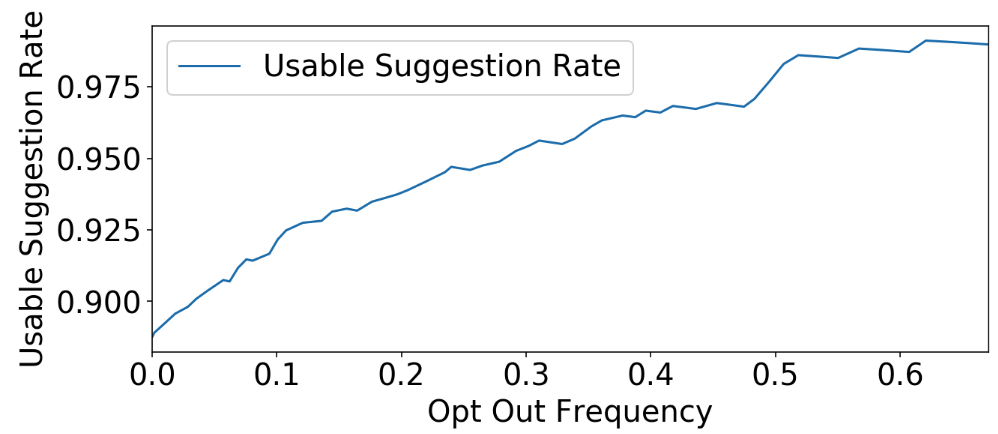}
\caption{The rate of bad suggested responses falls if we "opt-out", and don't suggest any response when the model's predicted probability is low. "Opt Out Frequency" measures how often the model chooses not to suggest a response, while "Usable Suggestion Rate" measures how often the suggested response is not worse than the doctor response observed in the data.}
\label{fig:opting}
\end{center}
\end{figure}

\textbf{How much history is useful?} We find, somewhat counterintuitively, that the ULMFit classifier does not benefit at all from using more than the last 6 turns of conversation history.  A table showing the accuracy using different amounts of history can be found in the appendix.
\newline \textbf{Well calibrated probabilities} Since the discriminative model is only generated on (context, response) pairs from a fixed bank of responses, it will occasionally see context that does not match any of the responses it is trained on. In these circumstances, it should not suggest a reply to the doctor. Figure \ref{fig:opting} shows that if we restrict our evaluations to the 50\% of situations where it is the most confident (as measured by the maximum predicted probability), the rate of bad suggested responses falls from 11\% to below 2\%.
\newline\textbf{Comparing different labeling procedures} We compare our 187 response group approach described in Section \ref{sec:Approach} with two other approaches: one using full automation with KMeans (897 clusters) and and another uses the full procedure with only 20 minutes of manual labeling (40 clusters). These approaches both generate roughly  35\% bad responses, according to expert evaluations, compared to 11\% for the 187 class approach that requires 3 hours of labeling.

\section{Conclusion}
In this work, we propose a classification model that leverages advances in pretraining techniques to generate useful responses in a wide variety of contexts while restricting generations to a fixed, easy to update set of high quality responses, thereby trading flexbility for control. We find that making this tradeoff also helps the average suggested response quality.

The key difficulty in this approach, and opportunity for future work is the grouping of response classes. We also intend to test whether the control for flexibility tradeoff provides similar quality improvements in other conversational domains.

\bibliographystyle{unsrtnat}
\bibliography{neurips_2019}
\section{Appendix}
\label{sec:appendix}

\subsection{Detailed Response Class Generation Procedure}
\begin{enumerate}
\item \textbf{Automatically Cluster Similar Responses}
\begin{enumerate}
  \item We lower case, remove patient and doctor identifying information, and remove punctuation from all responses seen in the data. We consider only the preprocessed responses $R$ that occur more than once, to make subsequent steps computationally cheaper.
  \item Estimating the similarity of every pair of responses is an O($\mathcal{R}^2)$ operation and most pairs are likely to have negligible similarity. Therefore, we restrict computing similarities of $r_i$
to only responses that are within a semantic neighborhood. More specifically, we encode each response as a vector using three pretrained sentence encoders: InferSent \citep{conneau-EtAl:2017:EMNLP2017}, the finetuned AWD-LSTM language model, the average Glove \citep{pennington2014glove} word vector for the response, and the TFIDF weighted average of the Glove vectors. For each encoder, we take the 10 nearest neighbors for each response. 
 $$
CandidatePairs =  \bigcup_{j \in encoders, i \in R}(i, KNN(enc_{j, -}, enc_{j, i}, 10))
$$
  \item  For each candidate pair, we run a supervised similarity model, BERT \citep{devlin2018bert} pretrained on Quora Question Pairs \citep{Wang_2018}, to predict the probability that each response pairs' members are semantically similar. We store the dissimilarity of each pair in a sparse distance matrix, with a distance of 1 (the maximum) if two responses were not blocked together. 
  $$
D_{i, j} =  
\begin{cases}
    (1 - ProbSimilar_{i,j}), & \text{if } (i,j) \in CandidatePairs\\
    1,              & \text{otherwise}
\end{cases}.
$$
  \item 
The last step is Agglomerative Clustering, on $D$,  where the distance between two responses is their predicted probability of dissimilarity or 1 if they were not generated as a candidate pair.  Merging requires complete linkage, which means that two clusters are only merged if all the responses in both clusters are >= 75\% similar to all responses in the other cluster. 
$$
Cluster_{i} = AgglomerativeClustering(D,  distanceThreshold=0.25)
$$
\end{enumerate}
\item \textbf{Manually Merge Clusters into Response Classes}
\begin{enumerate} 
\item Create dataset containing (centroid text, \# occurences of cluster constituents), sorted by \# occurrences, 
where the centroid is the most frequently occurring response in the cluster.
\item For each row in the dataset, the labeler decides whether the cluster centroid text belongs in any existing response class.
\begin{enumerate}
\item If centroid belongs in an existing group (most of the time): add the cluster to the existing response class.
\item Otherwise: create a new group with a memorable name, e.g “Greet + Pain Scale Question”.
\end{enumerate}
\end{enumerate}
We merge all responses that have the same impact on the user, and could therefore be used interchangeably. For example, even though “How long have you had the symptoms?” and “When did the symptoms start?” do not mean the same thing, they are both members of the same response class.

\end{enumerate}

\subsection{ULMFit Modifications}
Like the original work, we start with an AWD-LSTM language model pretrained on the wiki103 dataset \citep{merity2016pointer}, finetune the language model on our interaction history, and attach a classifier head to predict the response class given the concat pooled representation of final hidden state of the language model.

To accommodate larger batch size than the original work, which we found to help performance, we truncate context sequences to the last 304 tokens before passing them through the language model. This allows us to train with batches of 512 examples, and adjust the learning rate commensurately. 

To encode information about speaker changes, we insert two special tokens: one that indicates the beginning of the user’s turn and one that indicates the beginning of the doctor’s turn.

Finally, we add Label smoothing \citep{DBLP:journals/corr/PereyraTCKH17} with $t=0.1$ to the cross entropy loss function. Label smoothing smooths one-hot encoded classification labels 
towards $\frac{1}{numClasses}$, and reduces the impact of mislabeled examples on classification training.

\begin{figure}[!htbp]
\begin{center}
\includegraphics[scale=0.2]{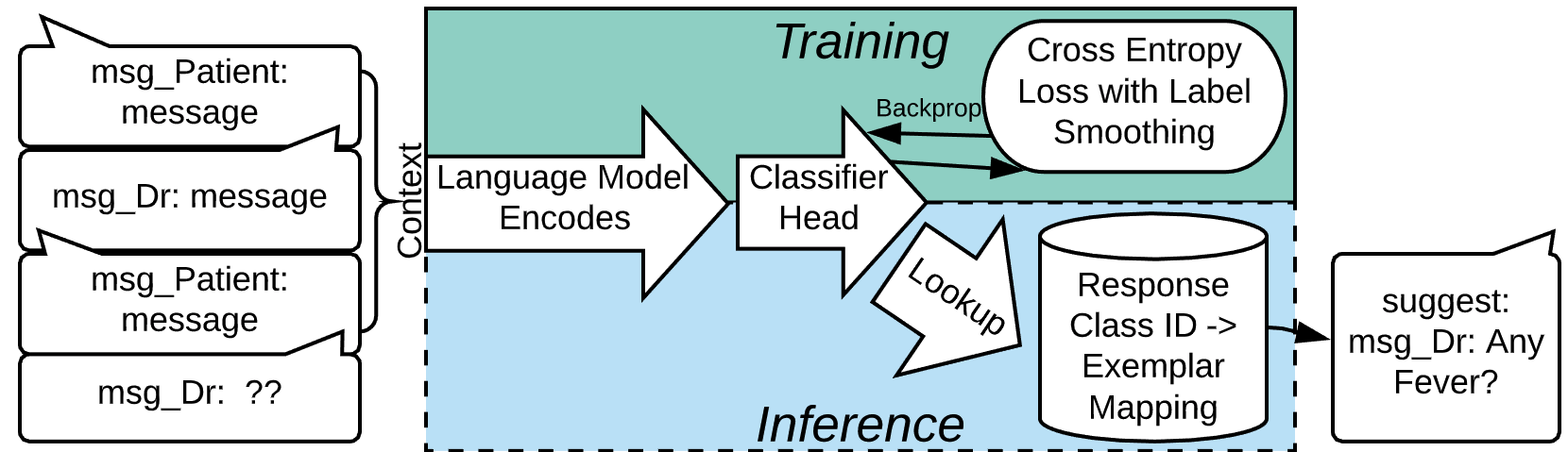}
\caption{Inference and training procedures, starting from a conversational context (left).}
\label{inference_diag}
\end{center}
\end{figure}



\textbf{Classification architecture comparison:} 
To facilitate comparison with the hierarchical encoding paradigm used by \citet{airbnb}, we tested two different architectures: hierarchical ULMFit (pretrained) and  hierarchical QRNN\footnote{All QRNN based experiments use random initialization, and 3 layers with hidden size 64.} (trained from scratch). In both settings, the higher level context RNN was a randomly initialized QRNN. We found that non-hierarchical ULMFit significantly outperformed its hierarchical counterpart while hierarchical and flat QRNN performed comparably. We attribute part of this discrepancy with previous work to the large variance in the length of each turn in our data. Turns vary from 2 to 304 tokens, after truncation, requiring models that consume 3D hierarchical encodings to consume large amounts of padding and smaller batch sizes. Hierarchical ULMFit on 8 turns could only be trained with batch size 32, while the non-hierarchical one fits 512 examples in each batch. To compare with \cite{hface}, we finetune a pretrained double headed transformer on our conversation data, discard the language modeling and multiple choice heads, and attach a one layer classification head that is trained until convergence. As shown in Table \ref{archtab}, this results in similar accuracy to the ULMFit architecture but is much more computationally expensive (10x train time, 20x slower inference).

\subsection{How much history is useful?}
\begin{table}[!htbp]
\centering
\begin{tabular}{@{}llllllllllll@{}}
\toprule
Max Turns of History   & 1      & 2      & 3      & 4      & 5      & 6      & 7      & All   \\ \midrule
Accuracy & 44.5\% & 53.3\% & 55.3\% & 56.7\% & 56.3\% & \textbf{57.7\%} & 57.4\% & 57.0\%  \\ \bottomrule
\end{tabular}
\caption{One turn is all messages sent consecutively by one conversation participant. Observations are truncated to the most recent $n$ turns.}
\label{ctxtab}
\end{table}

\subsection{Comparing Different Labeling Procedures}
\begin{table}[!htbp]
\centering
\begin{tabular}{@{}llll@{}}
\toprule
\# Classes & Train Examples & Bad Responses & Unique per 100 responses \\ \midrule
40$^ \dag$        & 19,300     & 38\%            & 17                       \\
187$^ \dag$       & 72,981     & 11\%            & 28                       \\
879$^ \phi$       & 86,941     & 34\%            & 49                       \\ \bottomrule
\end{tabular}

\caption{$^\dag$ Generated with process described in Section 2, including manual merge step. $^\phi$ Generated with KMeans and no manual merging or review. Bad responses percentage is calculated on 100 test set examples using the the manual evaluation process outlined above. Unique per 100 responses measures how many unique responses are generated per 100 conversation contexts, and is computed on 1000 test set suggestions.}
\label{rset_ablation}
\end{table}

\subsection{Opting out at different thresholds}
\end{document}